\documentclass[runningheads]{llncs}
\usepackage{booktabs}
\usepackage{multirow}
\usepackage{amsmath,amssymb}
\usepackage[T1]{fontenc}
\usepackage{graphicx,verbatim}
\begin{document}
\title{Dense Structural Priors for Sparse Functional Landmark Localization in Surgical Videos}
\titlerunning{Surgical Functional Landmark Localization}
%

\author{Chenyan Jing\inst{1} \and Hao Ding\inst{1} \and
Lalithkumar Seenivasan\inst{1} \and Jacob M. Delgado López\inst{1} \and
Mathias Unberath\inst{1}\thanks{Corresponding author}}

\authorrunning{C. Jing et al.}

\institute{Johns Hopkins University, Baltimore, MD, USA \\
\email{\{cjing5,unberath\}@jhu.edu}}
  
\maketitle              
\begin{abstract}
Vision foundation models such as SAM~3 can provide transferable object-level structure across diverse surgical video conditions, but segmentation outputs do not explicitly encode the action-conditioned semantics that define functional surgical landmarks. Estimating instrument extent and geometry differs from localizing the tip or anchor relevant to clipping, grasping, or dissecting. We investigate vision foundation model-enabled sparse action-aware landmark localization, using zero-shot, point-prompted structural masks to provide dense instrument-level context without manual pixel-level mask annotations. We propose a lightweight refinement framework that uses SAM~3 as a structural prior. A coarse multi-frame network predicts tip and anchor prompts, generating non-oracle masks that are fused with visual and heatmap features to refine functional landmark predictions. We compare direct mask-augmented supervision, prediction-derived mask-prior refinement, and auxiliary mask supervision to examine how vision foundation model-derived structure should enter a precision-oriented localization system. Experiments on 7,867 clips from 60 surgical videos spanning YouTube, Cholec80, HeiChole, SurgVU, and CRCD evaluate the approach under heterogeneous conditions. Without manual pixel-level mask annotations for training, the proposed model achieves overall F1 scores of 72.4\% for tip and 58.0\% for anchor localization. Ablations show that coarse-to-fine refinement provides a substantial performance gain, while prediction-derived structural priors provide additional improvement when incorporated as intermediate guidance rather than direct localization targets.

\end{abstract}

\section{Introduction}

Accurate localization of surgical tool landmarks is important for instrument pose estimation~\cite{Xu25SurgRIPE,Wu25SurgPose,Park24PrecisePose}, motion analysis and surgical skill assessment~\cite{Law17}, and computer-assisted interventions and surgical robotics~\cite{Nema22}. We focus on sparse tip and anchor landmarks for clipping, grasping, and dissecting actions, whose locations depend on the active tool and ongoing surgical action. Unlike generic instrument detection, these landmarks encode action-dependent functional locations rather than only instrument presence.

Localization is challenging under occlusion, smoke, specular reflection, motion blur, ambiguous boundaries, and multiple visible instruments~\cite{Nema22}. Vision foundation models such as SAM~2~\cite{Ravi24SAM2}, and SAM~3~\cite{carion2025sam3} can provide instrument-level masks from sparse prompts, capturing extent, shape, and coarse spatial continuity. Their medical and surgical adaptations have also shown promise for instrument segmentation~\cite{Ma24MedSAM,Wang23SAMRobotic,SurgicalSAM24,SurgicalDeSAM24}. Zero-shot SAM~3 masks are appealing for heterogeneous surgical videos, where pixel-level annotations are costly to obtain.

However, object-level masks do not directly identify the action-relevant tip or anchor. This creates a gap between the broad geometry provided by vision foundation models and the fine-grained functional precision required for landmark localization~\cite{Hariharan15Hypercolumns}. Although surgical instrument segmentation has been extensively studied in robotic and laparoscopic videos~\cite{Hong20CholecSeg8k,Allan19EndoVis,Allan20EndoVis,Zhao17Tracking,Rueckert23MISReview}, it remains unclear how foundation-model masks should be incorporated into heatmap-based landmark localization. Direct mask targets may bias predictions toward broad tool regions, whereas landmark prediction requires sharp, spatially specific, and action-relevant responses.

In this work, we formulate sparse functional landmark localization as a vision foundation model-enabled refinement problem. Coarse landmark predictions prompt SAM~3 to obtain non-oracle mask priors, which are fused with visual and heatmap features in a lightweight refinement network. We compare direct mask-augmented supervision, prediction-derived mask-prior refinement, and auxiliary mask supervision. Our results show that dense structural information is most effective as intermediate refinement guidance rather than a direct heatmap target.

\section{Method}
\begin{figure}[t]
    \centering
    \includegraphics[
        width=\linewidth,
        trim=0cm 0.4cm 0cm 0cm,
        clip
    ]{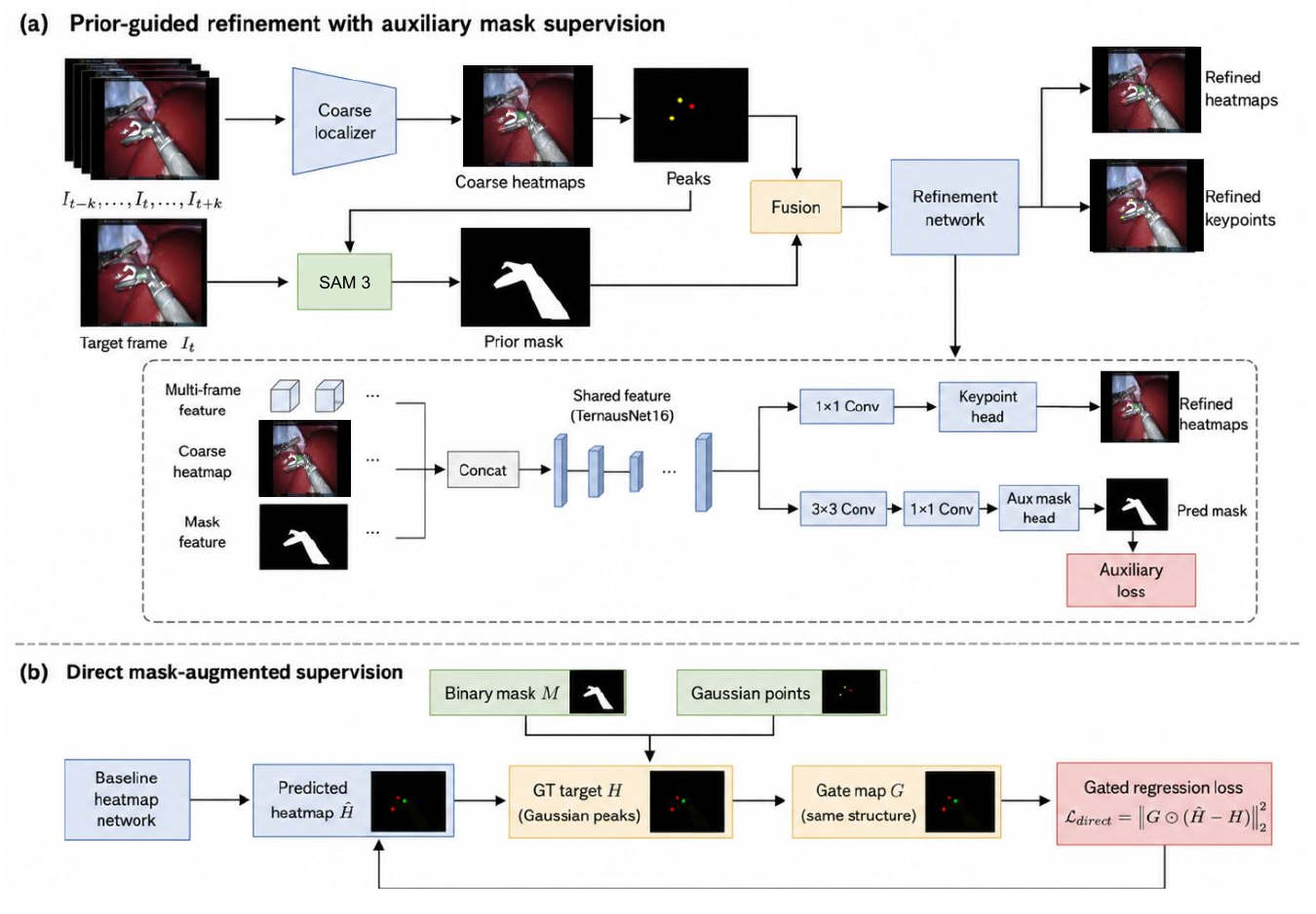}
\caption{Overview of the proposed framework and ablation settings. (a) A lightweight task-specific adapter uses coarse landmark predictions to prompt zero-shot SAM~3 and converts the resulting instrument-level structural prior into refined action-dependent landmark heatmaps. (b) Direct mask-augmented supervision uses dense structure as part of the heatmap target, serving as an ablation of target-level structural integration.}

    \label{fig:method}
\end{figure}

\subsection{Overview}
We formulate functional landmark localization as a coarse-to-fine prediction problem that combines temporal visual context with dense instrument-level structural guidance. All experiments use a multi-frame heatmap localization backbone adapted from MFCNet~\cite{ghanekar2025mfcnet}. For each target frame \(I_t\), the backbone takes an 8-frame temporal window
\[
\{I_{t-3}, I_{t-2}, I_{t-1}, I_t, I_{t+1}, I_{t+2}, I_{t+3}, I_{t+4}\}
\]
and predicts action-dependent tip and anchor heatmaps for \(I_t\). Consecutive target frames are processed using overlapping temporal windows, allowing neighboring frames to provide temporal context while localization remains defined on a single target frame. At video boundaries, unavailable context frames are padded by replicating the nearest available boundary frame. The main refinement framework is described in Sections~2.2 and~2.3, while direct mask-augmented supervision is introduced separately in Section~2.4. 

\subsection{Main Framework: Mask-Prior Refinement}

Our main framework uses SAM~3 as a zero-shot structural prior rather than as a final landmark predictor. The coarse network predicts target-frame tip and anchor heatmaps \(\hat{H}^{c}\), from which the decoded landmark locations are jointly used as positive point prompts for SAM~3. The resulting instrument mask \(M^{c}\) is prediction-derived and non-oracle, since no ground-truth landmark locations or manually annotated pixel-level masks are used to generate the structural prior during inference.

The refinement network concatenates the target-frame visual feature, coarse heatmap feature, and SAM~3-generated mask-prior feature along the channel dimension, and processes the fused representation to predict refined heatmaps:
\[
\hat{H}^{r}
=
F_{\text{refine}}(I_t, \hat{H}^{c}, M^c).
\]

\subsection{Auxiliary Supervision and Training Objective}

The refinement network can optionally include an auxiliary mask branch during training. As shown in Fig.~\ref{fig:method}(a), this branch is attached to the shared refinement feature after input fusion, rather than to the final keypoint prediction head. It predicts the SAM~3-generated mask prior \(M^c\), encouraging mask-shape information in the shared feature representation without requiring the final keypoint head to regress the broad mask region. The refinement model is trained using a keypoint heatmap loss, and when the auxiliary branch is included, it also incorporates an additional mask prediction loss:
\[
L
=
\left\|
\hat{H}^{r}-H^{gt}
\right\|_2^2
+
\lambda_{\text{aux}}
\left\|
\hat{M}-M^c
\right\|_2^2 .
\]
Here, \(H^{gt}\) denotes the Gaussian ground-truth heatmaps, \(\hat{M}\) is the predicted auxiliary mask, and \(\lambda_{\text{aux}}\) controls the auxiliary loss weight. For refinement without auxiliary supervision, \(\lambda_{\text{aux}}=0\). The direct mask-augmented ablation is trained separately using \(L_{\text{direct}}\).

\subsection{Ablation: Direct Mask-Augmented Supervision}

As an ablation, we study target-level structural integration using SAM~3 masks generated from ground-truth tip and anchor prompts. These masks are not manually annotated pixel-level segmentations. This oracle-prompted setting is used only for this ablation; the proposed framework instead prompts SAM~3 with coarse model predictions at inference time.

The baseline heatmap network is unchanged, while the Gaussian-only target is replaced by a mask-augmented target \(\tilde{H}\):
\[
\tilde{H}(x)
=
\begin{cases}
P(x), & x \in \Omega_{\text{kpt}},\\
\alpha, & M(x)=1 \ \text{and}\ x \notin \Omega_{\text{kpt}},\\
0, & M(x)=0,
\end{cases}
\]
where \(M\) is the SAM~3 mask generated from ground-truth landmark prompts, \(P\) denotes the Gaussian keypoint maps, \(\Omega_{\text{kpt}}\) denotes keypoint-centered regions, and \(\alpha=0.1\). This introduces weak structural guidance while retaining localized Gaussian targets.

We apply a gated regression loss:
\[
L_{\text{direct}}
=
\left\|
G \odot
\left(
\hat{H}^{d}-\tilde{H}
\right)
\right\|_2^2 ,
\]
where \(G\) assigns separate weights to keypoint, mask, and background regions, with the largest weights on keypoints.

\section{Experimental Setup}
\subsection{Dataset}
We evaluate an internally assembled surgical tool landmark localization dataset of 7,867 clips from 60 videos collected from YouTube, Cholec80~\cite{Twinanda17EndoNet}, HeiChole~\cite{Wagner23HeiChole}, SurgVU~\cite{Zia25SurgVU}, and CRCD~\cite{Oh24CRCD}. Frame-level annotations specify the active tool, action, tool type, and corresponding tip and anchor landmarks. Tip landmarks denote the distal functional endpoint(s) of the active instrument, while anchor landmarks provide proximal geometric references, such as an articulation center or predefined shaft-boundary points depending on instrument geometry. Each 8-frame input is supervised and evaluated using the annotation of its target frame. We evaluate clipping with clip appliers, grasping with graspers, and dissecting with hook instruments; cutting with scissors is excluded from evaluation due to limited data. Splits are performed at the surgical-video level to avoid clip-level leakage, with 6,403, 79, and 1,385 clips for training, validation, and testing, respectively. SAM~3 mask priors are generated zero-shot and require no manually annotated pixel-level instrument masks. Dataset composition is summarized in Table~\ref{tab:dataset_distribution}.

\begin{table}[t]
\centering
\footnotesize
\setlength{\tabcolsep}{6pt}
\caption{Dataset distribution by action and source (clip counts). Number of cases per source: YouTube~7, Cholec80~25, HeiChole~7, SurgVU~13, CRCD~8. The overall train/validation/test split is 6{,}403/79/1{,}385.}
\label{tab:dataset_distribution}
\begin{tabular}{lcccccc}
\toprule
Action & YouTube & Cholec80 & HeiChole & SurgVU & CRCD & Total\\
\midrule
Clip    & 36  & 157  & 57  & 35  & 0   & 285\\
Dissect & 611 & 3605 & 939 & 245 & 628 & 6028\\
Grasp   & 124 & 752  & 220 & 322 & 136 & 1554\\
\midrule
Total   & 771 & 4514 & 1216 & 602 & 764 & 7867\\
\bottomrule
\end{tabular}
\end{table}

\subsection{Training, Evaluation, and Compared Methods}
For external comparisons, we evaluate precision, recall, F1 score, and mean Euclidean localization error under a 15-pixel matching tolerance. We compare our approach with SimpleBaseline~\cite{Xiao18SimpleBaseline}, YOLOv8-pose~\cite{Yaseen24YOLOv8}, and RTMPose~\cite{Jiang23RTMPose}. Predictions associated with visually prominent but action-irrelevant instruments are treated as incorrect.

For evaluation, predicted and ground-truth landmarks are matched using a one-to-one matching protocol. A prediction is considered correct when it has the same landmark type and lies within 15 pixels of a ground-truth point. Unmatched predictions are counted as false positives, whereas unmatched ground-truth landmarks are counted as false negatives.

\section{Results and Discussion}

\subsection{Comparison with External Baselines}

Table~\ref{tab:external_comparison} compares our prior-guided refinement model with SimpleBaseline~\cite{Xiao18SimpleBaseline}, YOLOv8-pose~\cite{Yaseen24YOLOv8}, and RTMPose~\cite{Jiang23RTMPose}. 

Our method achieves the highest overall F1 score for both tip and anchor localization. It obtains 72.4\% F1 for tips and 58.0\% F1 for anchors, indicating improved overall detection performance across the evaluated actions. The method achieves the strongest results for tip localization in the clip and grasp settings, while YOLOv8-pose obtains the highest tip F1 for dissecting sequences.

For anchor localization, our method achieves the best overall F1 despite variation across actions. RTMPose produces slightly lower localization errors in the overall comparison, whereas the proposed method provides higher overall F1 scores for both landmark types.

\begin{table}[!t]
\centering
\footnotesize
\setlength{\tabcolsep}{2.0pt}
\renewcommand{\arraystretch}{1.06}
\caption{
External comparison at a 15-pixel matching tolerance.
Precision (P), recall (R), and F1 are reported in percentage;
L2 denotes Euclidean localization error in pixels.
Higher P/R/F1 and lower L2 are better.
}
\label{tab:external_comparison}
\begin{tabular}{@{}llcccccccc@{}}
\toprule
& & \multicolumn{4}{c}{Tip localization}
& \multicolumn{4}{c}{Anchor localization} \\
\cmidrule(lr){3-6}
\cmidrule(lr){7-10}
Model & Method
& P$\uparrow$ & R$\uparrow$ & F1$\uparrow$ & L2$\downarrow$
& P$\uparrow$ & R$\uparrow$ & F1$\uparrow$ & L2$\downarrow$ \\
\midrule

\multicolumn{10}{l}{\textit{Clip}} \\
SimpleBaseline & Heatmap
& 93.3 & 25.5 & 40.1 & 8.88
& 70.1 & 17.8 & 28.8 & 11.61 \\
YOLOv8-pose & Top-down
& 47.1 & 26.2 & 33.6 & 8.02
& 65.6 & 41.9 & 51.2 & \textbf{6.54} \\
RTMPose & Top-down
& 50.2 & 50.2 & 50.2 & 12.73
& 49.3 & 49.4 & 49.4 & 9.07 \\
Ours & Prior + aux refinement
& 77.9 & 48.4 & \textbf{59.7} & 9.05
& 61.5 & 51.1 & \textbf{55.8} & 10.19 \\
\midrule

\multicolumn{10}{l}{\textit{Grasp}} \\
SimpleBaseline & Heatmap
& 88.4 & 39.1 & 54.2 & 8.96
& 63.2 & 25.9 & 36.7 & 12.31 \\
YOLOv8-pose & Top-down
& 52.5 & 31.5 & 39.4 & 10.52
& 60.4 & 43.4 & \textbf{50.5} & 11.50 \\
RTMPose & Top-down
& 58.7 & 58.7 & 58.7 & 7.29
& 35.8 & 35.8 & 35.8 & \textbf{7.39} \\
Ours & Prior + aux refinement
& 80.9 & 49.7 & \textbf{61.6} & \textbf{7.23}
& 56.2 & 41.6 & 47.8 & 10.36 \\
\midrule

\multicolumn{10}{l}{\textit{Dissect}} \\
SimpleBaseline & Heatmap
& 46.4 & 26.4 & 33.6 & 14.97
& 78.2 & 39.9 & 52.8 & 10.49 \\
YOLOv8-pose & Top-down
& 84.7 & 72.2 & \textbf{78.0} & 10.04
& 52.6 & 45.2 & 48.6 & 12.02 \\
RTMPose & Top-down
& 69.3 & 68.8 & 69.1 & \textbf{8.90}
& 59.1 & 59.4 & \textbf{59.2} & \textbf{8.28} \\
Ours & Prior + aux refinement
& 81.1 & \textbf{72.6} & 76.6 & 10.47
& 67.6 & 52.2 & 58.9 & 8.49 \\
\midrule

\multicolumn{10}{l}{\textit{Overall}} \\
SimpleBaseline & Heatmap
& 53.1 & 28.5 & 37.1 & 13.38
& 77.3 & 39.4 & 52.2 & 10.58 \\
YOLOv8-pose & Top-down
& 77.4 & 60.1 & 67.7 & 10.05
& 53.3 & 45.0 & 48.8 & 11.89 \\
RTMPose & Top-down
& 65.5 & 65.2 & 65.4 & \textbf{8.70}
& 57.0 & \textbf{57.2} & 57.1 & \textbf{8.25} \\
Ours & Prior + aux refinement
& \textbf{81.0} & \textbf{65.5} & \textbf{72.4} & 9.80
& 66.6 & 51.3 & \textbf{58.0} & 8.64 \\
\bottomrule
\end{tabular}
\end{table}

\subsection{Bridging Structural Generalization and Functional Precision}

Table~\ref{tab:internal_ablation} examines how SAM~3-derived structural information should be incorporated into sparse, action-aware landmark localization. We compare direct mask-augmented targets with refinement-based use of prediction-derived mask priors and auxiliary mask supervision.

For tip localization, coarse-to-refine prediction improves overall F1 from 52.9\% to 68.5\%. Adding the SAM~3 mask prior further increases F1 to 71.6\%, and the full model with auxiliary supervision achieves the best result of 72.4\%. Direct mask-augmented supervision also improves tip localization to 68.7\%, indicating that structural cues are useful, but it remains below the prior-guided refinement variants.

For anchor localization, refinement improves F1 from 51.5\% to 56.4\%, while the mask prior further improves it to 57.9\%. The full model reaches the best anchor F1 of 58.0\% and reduces L2 error from 8.85 to 8.64 pixels compared with refinement plus prior alone. In contrast, direct mask-augmented supervision reaches 56.4\% F1 despite higher recall, suggesting that dense mask targets can bias predictions toward broad tool regions.

Overall, the results suggest that broad structural coverage and precise functional localization are better handled separately. Prediction-derived mask priors provide useful instrument-level context during refinement, while preserving sparse, action-dependent heatmap targets for final landmark prediction.

\begin{table}[!t]
\centering
\footnotesize
\setlength{\tabcolsep}{1.8pt}
\renewcommand{\arraystretch}{1.08}
\caption{Internal ablation of mask-guided refinement strategies.
Detection precision, recall, and F1 are evaluated at a 15-pixel matching threshold.
L2 denotes the average localization error in pixels.}
\label{tab:internal_ablation}

\begin{tabular}{llcccccccccc}
\toprule
\multirow{2}{*}{Method}
& \multirow{2}{*}{LM}
& \multicolumn{2}{c}{Clip}
& \multicolumn{2}{c}{Grasp}
& \multicolumn{2}{c}{Dissect}
& \multicolumn{4}{c}{Overall} \\
\cmidrule(lr){3-4}
\cmidrule(lr){5-6}
\cmidrule(lr){7-8}
\cmidrule(lr){9-12}
&
& F1$\uparrow$ & L2$\downarrow$
& F1$\uparrow$ & L2$\downarrow$
& F1$\uparrow$ & L2$\downarrow$
& P$\uparrow$ & R$\uparrow$ & F1$\uparrow$ & L2$\downarrow$ \\
\midrule

\multirow{2}{*}{Coarse}
& Tip
& 53.4 & 9.02
& 60.5 & 7.40
& 51.7 & 10.44
& 61.1 & 46.7 & 52.9 & 9.92 \\

& Anchor
& 45.9 & 9.30
& 46.5 & 10.49
& 51.8 & \textbf{8.26}
& 67.3 & 41.7 & 51.5 & \textbf{8.36} \\
\midrule

\multirow{2}{*}{Refine w/o prior}
& Tip
& 57.9 & 8.59
& 61.6 & 7.23
& 71.4 & 10.76
& 76.0 & 62.4 & 68.5 & 9.96 \\

& Anchor
& 54.0 & 11.17
& \textbf{47.9} & 10.39
& 57.2 & 8.29
& 63.1 & 50.9 & 56.4 & 8.49 \\
\midrule

\multirow{2}{*}{Direct mask-aug.}
& Tip
& 57.4 & 9.16
& \textbf{64.1} & 7.32
& 74.7 & \textbf{10.12}
& 72.9 & 65.0 & 68.7 & \textbf{8.95} \\

& Anchor
& 43.4 & 12.82
& 44.9 & 9.88
& \textbf{59.6} & 8.45
& 58.4 & \textbf{54.6} & 56.4 & 8.78 \\
\midrule

\multirow{2}{*}{Refine + prior}
& Tip
& 58.5 & \textbf{7.98}
& 61.6 & \textbf{7.21}
& 75.7 & 10.65
& 80.3 & 64.7 & 71.6 & 9.87 \\

& Anchor
& \textbf{63.8} & 10.69
& 47.5 & \textbf{10.16}
& 58.7 & 8.71
& 66.3 & 51.4 & 57.9 & 8.85 \\
\midrule

\multirow{2}{*}{Refine + prior + aux}
& Tip
& \textbf{59.7} & 9.05
& 61.6 & 7.23
& \textbf{76.6} & 10.47
& \textbf{81.0} & \textbf{65.5} & \textbf{72.4} & 9.80 \\

& Anchor
& 55.8 & 10.20
& 47.8 & 10.36
& 58.9 & 8.49
& \textbf{66.6} & 51.3 & \textbf{58.0} & 8.64 \\
\bottomrule
\end{tabular}
\end{table}

\subsection{Action- and Landmark-Dependent Analysis}
The effect of mask guidance varies across actions and landmark types. For clipping, refinement-level mask guidance is especially beneficial for anchor localization: the mask-prior refinement model reaches 63.8\% anchor F1, compared with 45.9\% for the coarse baseline and 43.4\% for direct mask-augmented supervision. For dissecting, the full model yields the strongest and most consistent performance, with 76.6\% tip F1 and 58.9\% anchor F1. In contrast, grasping shows weaker gains from mask guidance. One possible explanation is that grasping scenes often contain multiple visible instruments, while annotations are defined only for the active tool. This result highlights the distinction between object-level structural quality and action-dependent functional relevance. Even a plausible mask prior may be insufficient when it is not semantically aligned with the active instrument and surgical action, motivating the need for a task-specific adapter rather than direct reliance on segmentation output.

\begin{figure}[t]
    \centering
    \includegraphics[
        width=\linewidth,
        trim=0cm 35cm 0cm 4cm,
        clip
    ]{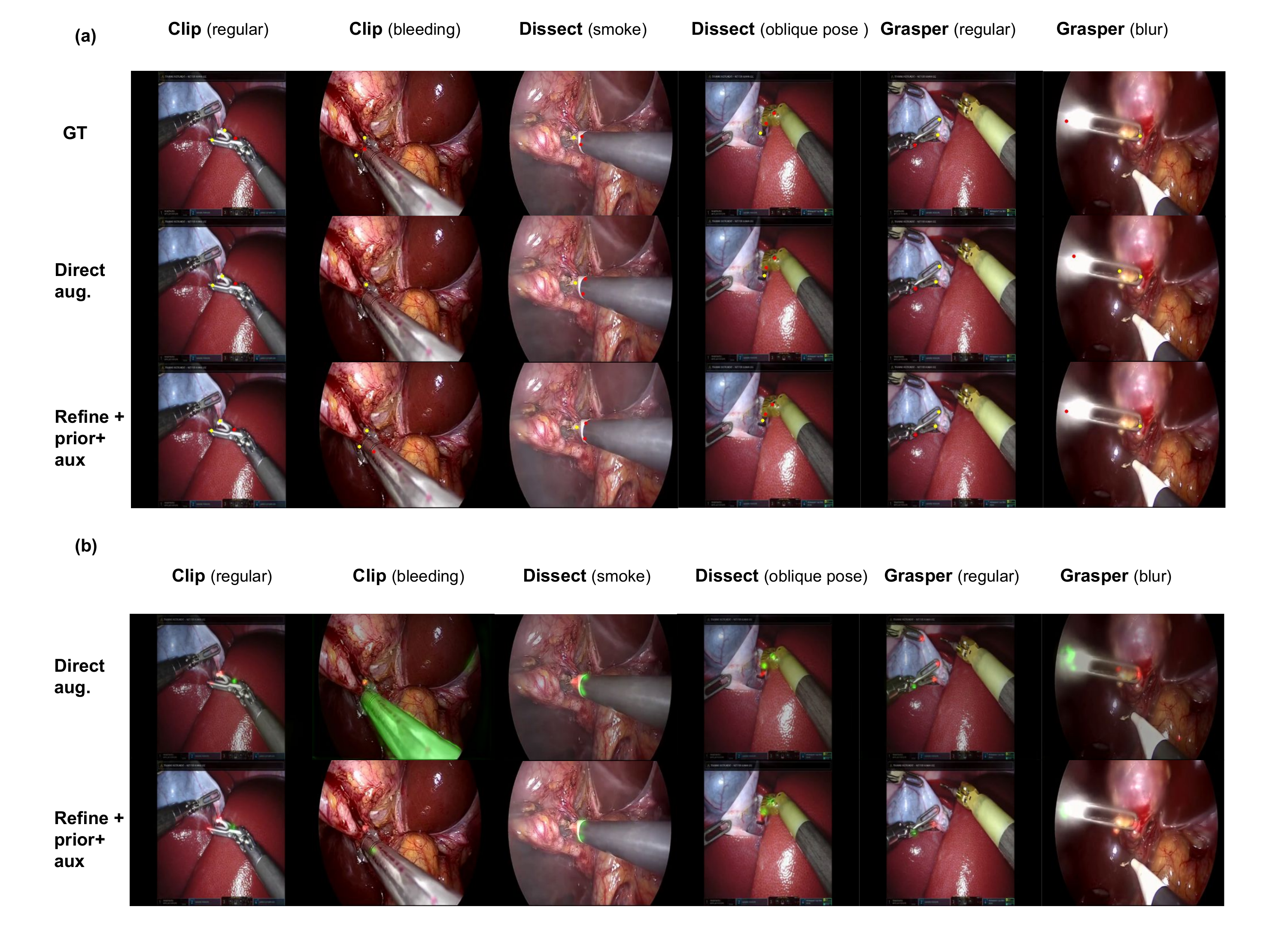}
    \caption{Qualitative comparison illustrating the tension between broad structural coverage and sharp functional localization. Direct mask-augmented supervision produces broader tool-region activations, whereas the proposed prior-guided adapter preserves localized responses around action-relevant landmarks while leveraging SAM~3-derived structural context.}

    \label{fig:qualitative}
\end{figure}

\subsection{Qualitative Results}
Figure~\ref{fig:qualitative} compares keypoint predictions and heatmap responses under representative surgical conditions, including bleeding, smoke, oblique tool pose, and motion blur. The examples highlight the gap between broad instrument-level coverage and precise action-dependent landmark localization. Direct mask-augmented supervision captures the visible instrument region but often produces broader activations because dense mask structure is imposed on the target representation. In contrast, prior-guided refinement retains sharper responses around action-relevant landmarks while leveraging mask-derived context. These examples support the use of dense structure as refinement guidance rather than a replacement for sparse keypoint heatmaps.

\section{Conclusion}
This work studies how dense mask-derived structural priors can be incorporated into sparse functional landmark localization for surgical tools. Our results show that mask information is useful, but its benefit depends on how it is introduced. Directly inserting masks into the heatmap target improves over the coarse baseline, yet the resulting broad mask plateau can bias learning toward tool regions rather than precise keypoint peaks. Rather than treating SAM~3 masks as direct landmark targets, our results show that zero-shot foundation-model geometry is most effective when adapted as intermediate context by a lightweight task-specific refinement network. This enables transferable instrument-level structure to support precise, action-dependent surgical landmark localization without requiring manual pixel-level mask annotations.

\bibliographystyle{splncs04}
\bibliography{report}

\end{document}